# Controllable Abstractive Summarization

**Angela Fan**  **David Grangier**  **Michael Auli**

Facebook AI Research
Menlo Park, California, USA

## Abstract

Current models for document summarization disregard user preferences such as the desired length, style, the entities that the user might be interested in, or how much of the document the user has already read. We present a neural summarization model with a simple but effective mechanism to enable users to specify these high level attributes in order to control the shape of the final summaries to better suit their needs. With user input, our system can produce high quality summaries that follow user preferences. Without user input, we set the control variables automatically – on the full text CNN-Dailymail dataset, we outperform state of the art abstractive systems (both in terms of F1-ROUGE1 40.38 vs. 39.53 F1-ROUGE and human evaluation).

## 1 Introduction

Summarization condenses a document into a short paragraph or a single sentence while retaining core information. Summarization algorithms are either extractive or abstractive. Extractive algorithms form summaries by pasting together relevant portions of the input, while abstractive algorithms may generate new text that is not present in the initial document (Das and Martins, 2007; Nenkova et al., 2011).

This work focuses on abstractive summarization and, in contrast to previous work, describes mechanisms that enable the reader to control important aspects of the generated summary. The reader can select the desired *length* of the summary depending on how detailed they would like the summary to be. The reader can require the text to focus on *entities* they have a particular interest in. We let the reader choose the *style* of the summary based on their favorite source of information, e.g., a particular news source. Finally, we allow the reader to specify that they only want to summarize a portion of the article, for example the *remaining* paragraphs they haven't read.

Our work builds upon sequence-to-sequence models (Sutskever et al., 2014; Bahdanau et al., 2015), which have been extensively applied to abstractive summarization (Rush et al., 2015; Chopra et al., 2016; Nallapati et al., 2016; See et al., 2017; Paulus et al., 2017). These conditional language models use an encoder network to build a representation of the input document and a decoder network to generate a summary by attending to the source representation (Bahdanau et al., 2015).

We introduce a straightforward and extensible controllable summarization model to enable personalized generation and fully leverage that automatic summaries are generated at the reader's request. We show that (1) our generated summaries follow the specified preferences and (2) these control variables guide the learning process and improve generation even when they are set automatically during inference. Our comparison with state-of-the-art models on the standard CNN/DailyMail benchmark (Nallapati et al., 2016), a multi-sentence summarization news corpus, highlights the advantage of our approach. On both the entity-anonymized (+0.76 F1-ROUGE1) and full text versions (+0.85 F1-ROUGE1) of the dataset, we outperform previous pointer-based models trained with maximum likelihood despite the relative simplicity of our model. Further, we demonstrate in a blind human evaluation study that our model generates summaries preferred by human readers.

## 2 User Controllable Summarization

We introduce our summarization model and describe the control variables users can modify.

### 2.1 Convolutional Sequence-to-Sequence

Our approach builds upon the convolutional model of Gehring et al. (2017). The encoder and decoder are deep convolutional networks (LeCun et al., 1990). Both start with a word embedding layer followed by alternating convolutions with Gated Linear Units (GLU) (Dauphin et al., 2017). The decoder is connected to the encoder through attention modules (Bahdanau et al., 2015) that performs a weighted sum of the encoder outputs. The weights are predicted from the current decoder states, allowing the decoder to emphasize the parts of the input document which are the most relevant for generating the next token. We use multi-hop attention, i.e. attention is applied at each layer of the decoder.

In addition to attending over encoder states (Bahdanau et al., 2015), we also use intra-attention in the decoder to enable the model to refer back to previously generated words. This allows the decoder to keep track of its progress and reduces the generation of repeated information (Vaswani et al., 2017; Paulus et al., 2017). To combine encoder and decoder attention, we alternate between each type of attention at every layer.

Much prior work on the CNN-Dailymail benchmark employed pointer networks to copy rare entities from the input (Nallapati et al., 2016), which introduces additional complexity to the model. Instead, we rely on sub-word tokenization and weight sharing. We show this simple approach is very effective. Specifically, we use byte-pair-encoding (BPE) for tokenization, a proven strategy that has been shown to improve the generation of proper nouns in translation (Sennrich et al., 2016b). We share the representation of the tokens in the encoder and decoder embeddings and in the last decoder layer.

### 2.2 Length-Constrained Summarization

Summarization allows a reader with limited time to quickly comprehend the essence of a document. Controlling summary length enables reading with different time budgets: a document might be summarized as a five-word headline, a single sentence or a paragraph, each providing more and more detail.

To enable the user to control length, we first quantize summary length into discrete bins, each representing a size range. Length bins are chosen so that they each contain roughly an equal number of training documents. We then expand the input vocabulary with special word types to indicate the length bin of the desired summary, which allows generation to be conditioned upon this discrete length variable. For training, we prepend the input of our summarizer with a marker that indicates the length of the ground-truth summary.

At test time, we control the length of generated text by prepending a particular length marker token. Our experiments (§5.2) provide quantitative and qualitative evidence that the model effectively uses this variable: output length is easily controlled by changing the length marker and supplying ground truth markers drastically improves summary quality. We compare our method to Kikuchi et al. (2016) and demonstrate that our straightforward length control strategy is more effective.

### 2.3 Entity-Centric Summarization

The reader might be interested in a document to learn about specific entities, such as people or locations. For example, a sports fan reading about a recent game might want to focus the summary on the performance of their favorite player. To enable entity-centric summaries, we first anonymize entities by replacing all occurrences of a given entity in a document by the same token. For training, we also anonymize the corresponding reference summary. For a (document, summary) pair, each entity is replaced with a token from the set (@entity0, ..., @entityN). This abstracts away the surface form, allowing our approach to scale to many entities and generalize to unseen ones.

We then express that an entity should be present in the generated summary by prepending the entity token to the input — prepending @entity3 expresses that the model should generate a summary where @entity3 is present. In effect, this instructs the model to focus on sentences that mention the marked entities.

At training time, we prepend each document with markers referring to an entity from the ground-truth summary. To ensure the entity request is informative, we provide an entity that is present in the ground-truth but not present in the summary generated by the baseline model. At test time, we may specify any entity marker that we wish the summary to contain. Our experiments (§5.2) evaluate the effect of prepending different markers to the input. We show that higher accuracy is achieved when we specify entities from the first few sentences of a document or if we supply markers taken from the reference summary to illustrate specific user preferences. We extend this approach to multiple entity markers and experiment with appending all ground-truth entities for training and provide all entities from Lead-3 at test time. We show that providing more entities improves summarization quality.

### 2.4 Source-Specific Summarization

Text sources such as newspapers and magazines often have specific style guidelines to provide a consistent experience. Readers are accustomed to the styles of their favorite sources. Therefore, we enable users to specify a preferred source style for a summary. Similar to length and entities, we introduce special marker tokens (@genSource0, ..., @genSourceN) to express source desiderata. For training, we preprend the input with the marker corresponding to the ground-truth source. At inference, we control the style of generated summary by prepending different markers. Our experiments (§4) evaluate whether providing the true source-style produces summaries that are closer to the reference summary. We additionally provide examples of distinct summaries resulting from changing source-style conditioning.

### 2.5 Remainder Summarization

Beyond reading summaries of full documents, readers may want the flexibility of only summarizing certain portions of a document. For example, a reader who has read the first few paragraphs would want a summary of the remaining text to cover what they missed.

Training and evaluating remainder summarization requires specific data, namely a dataset of full documents with position markers separating the already read portion from the remainder part along with the corresponding summaries. Such a dataset is not readily available and would be challenging to collect. To enable remainder summarization without such data, we align summaries to full documents. Our procedure matches each reference summary sentence to its best matching document sentence based on ROUGE-L. For any position in the document, we remove sentences aligned before this point from the full summary and consider this shorter summary as the summary of the remainder. In our experiment, we consider as read portions all article positions located at the middle of two alignment points, except for alignment points separated by less than 2 sentences.

We consider the following methods:
(1) **full summary baseline:** the baseline model predicts a full summary, disregarding the separation of the read portion from the remainder.
(2) **post-inference alignment:** a full summary is generated from the baseline model and the summary is shortened with our alignment procedure. The decoded summary sentences that align to the remainder portion compose the summary of the remainder.
(3) **remainder only**: the model is trained to map the document remainders to the remainder summaries on pre-aligned training data. This model is not given the read portion of the article.
(4) **read and remainder**: the model receives both read portion of the article and the remainder separated by a special token. It is trained to predict the remainder summary. We distinguish the read and remainder part of the article by using distinct sets of position embeddings.

We compare these methods in Section 4 and show the advantage of the model that receives both the user-read portion and the remainder of the document.

## 3 Related Work

### 3.1 Sequence-to-Sequence for Summarization

Automatic summarization has been an active research field for 60 years (Luhn, 1958). Extractive and abstractive methods have benefited from advances in natural language processing, pattern recognition, and machine learning (Nenkova et al., 2011). Recently, sequence-to-sequence neural networks (Sutskever et al., 2014) have been applied to abstractive summarization (Nallapati et al., 2016; See et al., 2017; Paulus et al., 2017) following their success in translation (Bahdanau et al., 2015; Luong et al., 2015b), parsing (Luong et al., 2015a) and image captioning (Vinyals et al., 2015b). Neural abstractive summarization has built upon advances from machine translation and related fields: attention (Bahdanau et al., 2015) enables generation to focus on parts of the source document while pointers (Vinyals et al., 2015a) help abstractive summarization to copy entities from the input (See et al., 2017; Paulus et al., 2017; Nallapati et al., 2016).

However, summarization also has distinct challenges. The generation of multi-sentence summaries differs from single sentence translation: left-to-right decoders need to be aware of their previous generation at a larger time scale, otherwise models tend to produce repeated text. To address this impediment, (See et al., 2017) introduce coverage modeling, (Paulus et al., 2017) propose intra-decoder attention, and (Suzuki and Nagata, 2017) equip the decoder with an estimator of unigram frequency. Previous work has also explored learning objectives: (Paulus et al., 2017) investigates replacing maximum likelihood training with Reinforcement Learning (RL) to optimize ROUGE, the most common automatic metric to assess summarization. Combining both strategies is found to perform best in human evaluations, as training with RL alone often produces non-grammatical text.

Our work builds upon prior research: like (Gehring et al., 2017), we rely on convolutional networks, which enable faster training. This contrasts with prior work using recurrent networks (Nallapati et al., 2016; See et al., 2017; Paulus et al., 2017). We borrow intra-attention from (Paulus et al., 2017) and expand it to multi-hop intra-attention inspired by multi-hop source attention from (Gehring et al., 2017). To facilitate copying input entities, we share the word representations between encoder and decoder (Paulus et al., 2017), and also rely on BPE tokenization (Sennrich et al., 2016b). This combination allows us to forgo an additional pointer mechanism unlike (Paulus et al., 2017; See et al., 2017; Nallapati et al., 2016). Unlike (Paulus et al., 2017), we did not explore training objectives and maximized the likelihood of the training summaries given the source document. Our model is amenable to RL, but this aspect is largely orthogonal to our main goal, i.e. controllable summarization.

## 3.2 Controllable Text Generation

Text generation is an established research area (McKeown, 1992). The field follows recent advances in generative models, such as the introduction of variational auto-encoders (Kingma and Welling, 2013) and adversarial networks (Goodfellow et al., 2014). This is exemplified by work focusing on natural language generation such as (Bowman et al., 2016; Yu et al., 2017; Zhao et al., 2017; Rajeswar et al., 2017).

Building upon unconditioned generation, controllable generation is an emerging research field. Research in computer vision includes style transfer (Gatys et al., 2015) or controllable image generation (Lample et al., 2017). Text generation work focuses on controlling tense or sentiment with variational auto-encoders (Hu et al., 2017). Shen et al. (2017) relies on adversarial training for manipulating sentence sentiment and Sennrich et al. (2016a) propose using side constraints for polite neural machine translation models. Takeno et al. (2017) extend the side constraints to control further aspects of translation output, such as length. Others have worked on style, for example Ficler and Goldberg (2017) propose using a conditional language model to generate text with stylistic requirements and Kobus et al. (2017) propose using tokens and additional features to translate text in different domains. Filippova (2017) proposes controlling length for generating answers in a question answering task. Kikuchi et al. (2016) explores length control for sentence compression using decoding-time restrictions and training-time length token embeddings.

Motivated by simplicity, our work relies on conditional language modeling and does not require adversarial training, latent variable models such as variational auto-encoders, or pointer networks. While latent variable models are popular for the generation of continuous outputs such as images, (conditional) language models are flexible enough to capture the multimodal nature of the data. We leave the assessment of how additional latent variables might improve upon our results to future work.

## 4 Experimental Setup

**Dataset**: We use the CNN-Dailymail dataset (Hermann et al., 2015; Nallapati et al., 2016). It consists of news articles along with multi-sentence summaries, with a total of 287k train, 13k valid and 11k test articles. On average, the articles are 758 token long, and the summaries are 55 token long. Most of our experiments are performed with articles truncated at 400 tokens, as suggested by (See et al., 2017). We evaluate on two versions of the data: the entity anonymized version (Hermann et al., 2015; Nallapati et al., 2016; Paulus et al., 2017) and the full text version (See et al., 2017). We use BPE with 30K types (Sennrich et al., 2016b) for most experiments. For non-BPE models, input and output vocabularies have resp. 47k and 21k word types, corresponding to types with more than 20 train occurrences.

Further, we compare length control with (Kikuchi et al., 2016) on DUC-2004 single-sentence summarization task. We train on English Gigaword following the protocol of Rush et al. (2015). The data consist of 3.6 million pairs (first sentence, headline of news articles). Following (Kikuchi et al., 2016), we evaluate on the 500 documents in the DUC2004 task-1. We use a source and target vocabulary of 30k words.

**Architecture, Training, and Generation**: We implement models with the fairseq library[1]. For CNN-Dailymail, our model has 8 layers in the encoder and decoder, each with kernel width 3. We use 512 hidden units for each layer, embeddings of size 340, and dropout 0.2. For DUC, we have 6 layers in the encoder and decoder with 256 hidden units.

Similar to Gehring et al. (2017), we train using Nesterov accelerated gradient method (Sutskever et al., 2013) with gradient clipping 0.1 (Pascanu et al., 2013), momentum 0.99, and learning rate 0.2. We reduce the learning rate by an order of magnitude when the validation perplexity ceases to improve, and end training when the learning rate drops below $10^{-5}$. Summaries are generated using beam search with beam size 5. To avoid repetition, we prevent the decoder from generating the same trigram more than once, following Paulus et al. (2017).

---

[1] github.com/facebookresearch/fairseq

**Evaluation**: On the CNN-Dailymail benchmark, our **automatic evaluation** reports F1-ROUGE scores for ROUGE-1, ROUGE-2, and ROUGE-L (Lin, 2004). We compare to existing abstractive baselines (Nallapati et al., 2016; See et al., 2017; Paulus et al., 2017). We also compare with Lead-3 which selects the first three sentences of the article as the summary. Note that, although simple, this baseline is not outperformed by all models.

For **human evaluation**, we conduct a human evaluation study using Amazon Mechanical Turk and the test set generation output of See et al. (2017). 500 articles from the test set were randomly selected and evaluated by 5 raters. The raters were presented with the first 400 words of each news article and asked to select the summarization output they preferred.

For the DUC-2004, we report recall ROUGE for ROUGE-1, ROUGE-2, and ROUGE-L at 30, 50, and 75 byte lengths following Kikuchi et al. (2016).

## 5 Results

We evaluate the design choices of our model and the impact of manipulating the control variables. We analyze the performance of the remainder summarization task and demonstrate the advantage of modeling both the read and remainder portions of the document.

### 5.1 Convolutional Summarization

Table 1 details the effect of our design choices for our baseline. Adding a constraint to avoid repeated trigrams at generation time improves F1-ROUGE1 by +2.86. Adding intra-attention to enable the model to examine past generations over long distances improves the accuracy obtained with the trigram constraint by a further 0.51 F1-ROUGE1. The modest improvement is likely because the two features address a similar problem of avoiding repeated generations. Switching tokenization from word to BPE gives another +0.79 F1-ROUGE1. BPE improves the ability to copy proper nouns and rare inflections, both of which are difficult to model in word-based vocabularies. This agrees with translation results (Sennrich et al., 2016b). Lastly, we find tuning the min/max length on the validation set and applying the constraints to the test set improves F1-ROUGE1 by 0.25.

| Model | ROUGE | | |
|---|---|---|---|
| | 1 | 2 | L |
| fairseq | 33.32 | 12.64 | 30.57 |
| + trigram decoding | 36.18 | 14.10 | 33.18 |
| + intra-attention | 36.69 | 14.28 | 33.47 |
| + BPE | 37.48 | 15.12 | 34.16 |
| + tuning min/max len | 37.73 | 15.03 | 34.49 |

Table 1: Baseline without control variables. Each row add a feature on top of the previous row features.

| Model | ROUGE | | |
|---|---|---|---|
| | 1 | 2 | L |
| baseline, no control | 37.73 | 15.03 | 34.49 |
| Length constraint | 39.16 | 15.54 | 35.94 |
| Entity centric | 38.17 | 15.16 | 34.92 |
| Source specific | 37.68 | 15.16 | 34.40 |
| Length+Entity+Source | 39.61 | 15.83 | 36.48 |

Table 2: Summarization with oracle control to simulate user preference.

| Model | ROUGE | | |
|---|---|---|---|
| | 1 | 2 | L |
| **Lead-3** | | | |
| Nallapati et al. (2017) | **39.2** | **15.7** | 35.5 |
| **Maximum Likelihood** | | | |
| Nallapati et al. (2016) | 35.46 | 13.30 | 32.65 |
| Paulus et al. (2017) | 37.86 | 14.69 | 34.99 |
| Paulus et al. + intra-attn | 38.30 | 14.81 | 35.49 |
| fairseq no control (ours) | 37.48 | 15.12 | 34.16 |
| + fixed control | **38.68** | **15.40** | 35.47 |
| + Lead-3 ent | **39.06** | **15.38** | **35.77** |
| **Reinforcement Learning** | | | |
| Paulus et al. (2017) | **39.87** | **15.82** | **36.90** |

Table 3: Fixed control variables on entity-anonymized text. Even with fixed variables, the controllable model improves ROUGE compared to ML alternatives.

### 5.2 Controllable Summarization

Our summarizer lets users control the length of the generated summary, entities on which it focuses on, and source style it imitates (see §2). We first evaluate the effect of providing the oracle reference variables at decoding time. This simulates a user setting their preferences to specific values. We then assess the effect of providing non-reference control variables.

Table 2 reports our results for each variable and their combined effect. All control variables improve the summary quality, but length control has the most impact, followed by entity control and source style. Further, the advantages of each control variable cumulatively produce an even stronger summary: we obtain +2.2 F1-ROUGE1 when combining control variables.

| Model | ROUGE | | |
|---|---|---|---|
| | 1 | 2 | L |
| **Lead-3** | **40.34** | **17.70** | **36.57** |
| **Maximum Likelihood** | | | |
| See et al. (2017) | 39.53 | 17.28 | 36.38 |
| fairseq no control (ours) | 38.23 | 16.68 | 34.77 |
| + fixed control | **39.75** | 17.29 | **36.54** |
| + Lead-3 ent | **40.38** | 17.44 | **37.15** |

Table 4: Summarization with fixed control variables on original text. Even with a fixed setting, the controlled summarization model improves ROUGE.

**Length control** improves accuracy by 1.68 F1-ROUGE1 (Table 2). This improvement is due to two effects: length mismatch is heavily penalized by F1-ROUGE. Moreover, the baseline struggles at predicting correct lengths. The latter is due to large uncertainty in summary length, i.e. even humans have difficulty predicting the correct length.

Figure 1 reports the average summary length when decoding all examples in the test set using each of the 10 possible length markers. The model is shown to respect length markers. Table 8 demonstrates the effect of the length marker on a specific example.

**Entity control** has less impact on ROUGE compared to length control at +0.69 vs. +1.68 F1-ROUGE1 (Table 2). This is mainly because our summaries often already contain most entities from the ground-truth without the need for additional instruction. Table 6 further analyzes entity control for 100 test documents. We decode repeatedly requiring each entity from lead-3. We then repeat the experiment with each entity from the full article. We report how often the entity-centric model generates a summary that actually contains the requested entity. For Lead-3 entities, the model mentions the requested entity $61\%$ of the time, while for all entities from the input, the model mentions required entities $34\%$ of the time. In both settings, these rates are much higher than the baseline. The model has difficulty generating summaries with entities which are unlikely to appear in the human references, e.g. unimportant entities far from the beginning of the article.

**Source-style control** is the least impactful control in terms of ROUGE, we report +0.2 F1-ROUGE1 in Table 2. Changing the source style variable changes the summary as shown in Table 8. Generally, we observe that generated summaries in the Dailymail-style are more repetitive and slightly longer than the CNN-style summaries. This matches the differences between the two sources in the reference text. The impact of style requests might be greater with a richer set of styles — in future work, we plan to evaluate on datasets where varied styles are available.

### 5.3 Summarization with Automatic Control

Our primary objective is to allow readers to control the attributes of generated summaries. However, we can also set the control variables automatically in absence of reader desiderata. For length and source-style, we set the variable to a constant value that maximizes ROUGE on the validation set. For entity control, we randomly sample an entity that appears in lead-3 and provide it as the entity of interest.

Table 3 reports results on the entity-anonymized version of the dataset like (Nallapati et al., 2016; Paulus et al., 2017) and Table 4 reports results on the full text data like (See et al., 2017). In both cases, our method is advantageous over alternatives. Further, providing all of the entities at training time and only lead-3 entities at test time improves quality. On the original text, we report 40.38 F1-ROUGE1 as opposed to 39.53 for (See et al., 2017). On the entity-anonymized text, we report 39.06 F1-ROUGE1 as opposed to 38.30 for the best maximum likelihood setting of (Paulus et al., 2017). We hypothesize that providing all lead-3 entities encourages copying from lead-3. Our model does not outperform the reinforcement learning model of (Paulus et al., 2017) which optimizes ROUGE. However, training objectives are orthogonal to our work on control variables and we expect reinforcement learning to equally benefit our model.

Table 5 compares results on DUC-2004 to the best method presented by (Kikuchi et al., 2016). We find that adding length embedding improves the ROUGE-1 and ROUGE-L scores for 30, 50, and 75 byte evaluation. Notably, ROUGE improves more for shorter text evaluation, likely because requesting a shorter document allows the model to plan its generation. Comparing to Kikuchi et al. (2016), our results are stronger while our method is very simple – (Kikuchi et al., 2016) explore embedding the remaining length at each timestep during decoding and creating a separate memory cell to control length. In contrast, we simply provide the desired length as a special token and show this simple approach is effective. Lastly, we note that length-control has less effect on DUC-2004 compared to CNN-Dailymail since truncated recall-ROUGE evaluation does not penalize length mismatch strongly.

| Model | 30 byte | | | 50 byte | | | 75 byte | | |
|---|---|---|---|---|---|---|---|---|---|
| | 1 | 2 | L | 1 | 2 | L | 1 | 2 | L |
| $LenInit_{(0,L)}$ (Kikuchi et al., 2016) | 14.31 | 3.27 | 13.19 | 20.87 | 6.16 | 19.00 | 25.87 | 8.27 | 23.24 |
| Baseline without control | 21.47 | **7.63** | 20.71 | 25.07 | **8.49** | 22.97 | 29.88 | **10.37** | 26.29 |
| + fixed length (ours) | **21.81** | 7.51 | **21.05** | **25.39** | 8.38 | **23.37** | **30.00** | 10.27 | **26.43** |

Table 5: ROUGE for fixed length control variable on DUC-2004 Task 1.

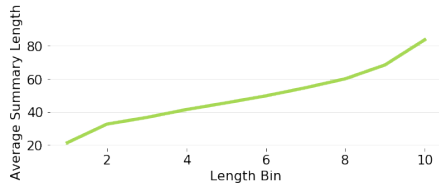

Figure 1: Length control vs summary length. Length control can take 10 discrete values.

| | Baseline | Entity-centric |
|---|---|---|
| Lead-3 | 15.28 | **61.15** |
| Full input | 7.64 | **33.76** |

Table 6: Fraction of requested entity actually occurring in decoded summaries. Entities originate either from lead-3 or from the full document.

Overall, the improvements from automatic control show that a better model can be obtained by providing additional information during training – we present the first model trained with maximum likelihood to match the strong Lead-3 baseline. When the model is not required to predict the summary length or the entities of interest, it can assign more capacity to generating text conditioned on these variables. This is particularly useful for variables which are hard to predict from the input due to intrinsic uncertainty like length. In subsequent work, we plan to explore architectures to explicitly divide the prediction of control variables and sequence-to-sequence mapping.

| Model | ROUGE | | |
|---|---|---|---|
| | 1 | 2 | L |
| Full summary | 28.12 | 9.46 | 18.81 |
| Post-inference align. | 27.13 | 7.68 | 27.45 |
| Remainder only | 30.30 | 11.44 | 27.46 |
| Read + remainder | 30.54 | **11.60** | 27.67 |
| Read + rem. + length | **30.70** | 11.52 | **27.78** |

Table 7: Remainder Summarization.

### 5.4 Remainder Summarization

Summarizing the remainder of an article helps the reader to quickly grasp what they have not yet read. Table 7 presents our results relying on aligned data introduced in §2.5. Generally, this task is more difficult than summarizing the entire article. First, the length of both read portions and summaries varies greatly. It is difficult for the model to distinguish information specific to the remaining portion of the document from the general point of the article. Despite this, when models trained on summarizing the remainder are tasked with summarizing only full documents, the performance is not much worse (37.02 F1-ROUGE1 compared to 37.73 F1-ROUGE1 of the baseline in Table 1).

Our baseline always presents the full summary, regardless of the portion of the article presented as input. It achieves an F1-ROUGE1 score of 28.12. Among our three proposed methods, forming the remainder summaries post-inference performs poorly as it depends largely on alignment quality. The news articles are repetitive, so one summary sentence can align to multiple locations in the source. Training the model to perform remainder summarization significantly improves our results. Models that receive only the remainder and produce a summary achieve F1-ROUGE1 of 30.30, while models that receive both the read and remainder portions are slightly better (F1-ROUGE1 30.54). We hypothesize that presenting the read portion of the article improves the quality as the model can focus on the new information in the remainder. An explicit method for eliminating redundancy between the read and the remainder is relevant future work.

Remainder summary length is particularly difficult to predict. We therefore rely on length control: we split the validation dataset into 10 partitions based on how far in the article the remainder begins and determine the best length setting for each partition. We decode the test data with this setting which provides an additional improvement, 30.70 F1-ROUGE1. However, partitioning is not an accurate length model and we hypothesize that length control could provide a greater improvement with a better model.

**a. Summary with Length Control**

**Requesting Length 2:** @entity0 [Easter] is over for the wild rabbits of greater @entity2 [Sydney] as councils and parks prepare another attempt to kill them off with a deadly virus. It comes after over 30 government bodies scattered carrots laced with calicivirus.

**Requesting Length 6:** @entity0 [Easter] is over for the wild rabbits of greater @entity2 [Sydney] as councils and parks prepare another attempt to kill them off with a deadly virus. This year, because of really high summer rainfall - which led to great food availability - there has been a big surge in the rabbit population in @entity2 [Sydney].

**Requesting Length 10:** @entity0 [Easter] is over for the wild rabbits of greater @entity2 [Sydney] as councils and parks prepare another attempt to kill them off with strategically placed carrots that have been laced with a deadly virus. This year,because of really high summer rainfall - which led to great food availability - there has been a big surge in the rabbit population in @entity2 [Sydney]. It comes after over 30 government bodies scattered carrots laced with calicivirus around public areas in March.

**b. Summary with Entity Control**                                      *blue highlights requested entity*

**Requesting @entity17 [Route 5]:** @entity1 [Linda MacDonald], 55 , was arrested for driving under the influence of alcohol Monday night in @entity4 [Dummerston], @entity5 [Vermont]. Police say the woman from @entity15 [Shelburne], @entity16 [Massachusetts] was driving drunk around 10:30pm when she ran off **@entity17 [Route 5]** in @entity4 [Dummerston].

**Requesting @entity20 [MacDonald]:** @entity1 [Linda MacDonald], 55 , was arrested for driving under the influence of alcohol Monday night in @entity4 [Dummerston], @entity5 [Vermont]. **@entity20 [MacDonald]** told officers that she crashed while talking on the phone and trying to take directions down on a legal note pad in her car. But when officers smelled alcohol on **@entity20 [MacDonald]**, they administered a breathalyzer test and she posted a .10 blood-alcohol content.

**c. Summary with Source-Style Control**                                *blue highlights different text*

**Requesting CNN-Style:** Officer @entity6 [Jared Forsyth], 33, had been a member of the @entity7 [Ocala Police Department] since 2012. He was wearing bulletproof vest, but round entered in his arm and went through his chest. @entity6 [Jared Forsyth] was rushed to hospital in critical condition.

**Requesting DailyMail-Style:** Officer @entity6 [Jared Forsyth], 33, had been a member of the @entity7 [Ocala Police Department] since 2012. He was rushed to @entity26 [Ocala Regional Medical Center] in critical condition and was taken into surgery. Police say the incident occurred about 3.30pm at a gun range at the @entity13 [Lowell Correctional Institution].

**d. Remainder Summary**

**Full Article:** @entity4 [Harry Potter] star says he has no plans to fritter his cash away on fast cars, drink and celebrity parties. @entity3 [Daniel Radcliffe]'s earnings from the first five @entity4 [Harry Potter] films have been held in a trust fund which he has not been able to touch.

**After 8 sentences:** He'll be able to gamble in a casino, buy a drink in a pub or see the horror film. @entity3 [Daniel Radcliffe]'s earnings from first five @entity4 [Harry Potter] films have been held in trust fund .

**After 12 sentences:** @entity3 [Daniel Radcliffe]'s earnings from first five @entity4 [Harry Potter] films have been held in trust fund .

Table 8: Summaries with various settings for user control variables and remainder summarization.

|  | ROUGE1 | Human Pref. |
| --- | --- | --- |
| (See et al., 2017) | 39.53 | 41.04% |
| fixed ctrl+Lead-3 ent. | **40.38** | **58.99%** |

Table 9: Human evaluation: 59% of ratings prefer our summaries (500 CNN-DM test articles, 5 raters each).

## 5.5 Human Evaluation

Our study compares summarization with fixed value control variables on full text CNN-Dailymail with (See et al., 2017). Table 9 shows that human raters prefer our model about 59% of the time based 2.5k judgments. Our model can therefore improve summary quality in a discernible way. As an aside, we find that ROUGE and ratings agree in two-thirds of the cases, where at least four out of five humans agree.

## 6 Conclusion

We proposed a controllable summarization model to allow users to define high-level attributes of generated summaries, such as length, source-style, entities of interest, and summarizing only remaining portions of a document. We simulate user preferences for these variables by setting them to oracle values and show large ROUGE gains. The control variables are effective without user input which we demonstrate by assigning them fixed values tuned on a held-out set. This outperforms comparable state of the art summarization models for both ROUGE and human evaluation.

**Acknowledgments** We thank Yann Dauphin for helpful discussions. We thanks Abigail See for sharing her summaries and Romain Paulus for clarifying their setup. We thank Jonas Gehring and Denis Yarats for writing the fairseq toolkit used for these experiments.